\documentclass[conference]{IEEEtran}
\IEEEoverridecommandlockouts
\usepackage{cite}
\usepackage{amsmath,amssymb,amsfonts}
\usepackage{algorithmic}
\usepackage{graphicx}
\usepackage{textcomp}
\usepackage{xcolor}
\def\BibTeX{{\rm B\kern-.05em{\sc i\kern-.025em b}\kern-.08em
    T\kern-.1667em\lower.7ex\hbox{E}\kern-.125emX}}
\usepackage{times}
 
\usepackage{soul}
 \usepackage{url}
\usepackage[hidelinks]{hyperref}
\usepackage[utf8]{inputenc}
\usepackage[small]{caption}
\usepackage{graphicx}
\usepackage{amsmath}
\usepackage{booktabs}
\usepackage{myformat}
\usepackage{tikz}  
\begin{document}    

\newcommand{\refeq}[1]{{Eq.~(\ref{#1})}}
\newcommand{\refeqs}[1]{{Eqs.~(\ref{#1})}}
\newcommand{\reffig}[1]{{Figure~\ref{#1}}}
\newcommand{\reftab}[1]{{Table~\ref{#1}}}
\newcommand{\reffigs}[1]{{Figures~\ref{#1}}}
\newcommand{\refsec}[1]{{Section~\ref{#1}}}

\newif\ifblacktext
\blacktextfalse

\ifblacktext
\newcommand{\sbs}[1]{\bf SBS: #1} 
\newcommand{\spf}[1]{\bf SPF: #1} 
\newcommand{\zlw}[1]{\bf ZLW: #1} 
\else
\newcommand{\sbs}[1]{\textcolor{orange}{SBS: #1}} 
\newcommand{\spf}[1]{\textcolor{blue}{SPF: #1}} 
\newcommand{\zlw}[1]{\textcolor{teal}{ZLW: #1}} 
\fi

\newcommand{\figscale}{0.75}
\title{Spikemax: Spike-based Loss Methods for Classification\\
}
\author{

\IEEEauthorblockN{Sumit Bam Shrestha$^{1*}$\thanks{$^{*}$Corresponding Author}, Longwei Zhu$^1$, Pengfei Sun$^{1, 2}$}
\IEEEauthorblockA{\textit{$^1$Institute for Infocomm Research, Agency for Science, Technology and Research (A*STAR), Singapore }\\ 
\textit{$^2$Ghent University, Belgium} \\
Email:  \{sumit\_bam@i2r.a-star.edu.sg, wayne\_zhu@i2r.a-star.edu.sg, pengfei.sun@ugent.be\}}\\
}

\maketitle

\begin{abstract}
Spiking Neural Networks~(SNNs) are a promising research paradigm for low power edge-based computing. Recent works in SNN backpropagation has enabled training of SNNs for practical tasks. However, since spikes are binary events in time, standard loss formulations are not directly compatible with spike output. As a result, current works are limited to using mean-squared loss of spike count. In this paper, we formulate the output probability interpretation from the spike count measure and introduce spike-based negative log-likelihood measure which are more suited for classification tasks especially in terms of the energy efficiency and inference latency. We compare our loss measures with other existing alternatives and evaluate using classification performances on three neuromorphic benchmark datasets: NMNIST, DVS Gesture and N-TIDIGITS18. In addition, we demonstrate state of the art performances on these datasets, achieving faster inference speed and less energy consumption.
\end{abstract}


\section{Introduction}

Error backpropagation is the key technology that propels the current deep learning revolution. The basic strategy for training these Artificial Neural Networks~(ANNs) is to compute the gradient of a loss measure of the output of the network and its desired target, and use gradient descent steps to arrive at an optimal/sub-optimal set of parameters. It has been very successful in various applications ranging from image classification, object recognition, object tracking, signal processing, natural language processing etc. and many more.

Spiking Neural Networks~(SNNs) have garnered increased interest in recent times as a promising option for extremely low power edge-based computing devices with the development of neuromorphic hardware \cite{Merolla2014,Furber2014,Davies2018,Neckar2018,Pei2019}. They are a more biologically plausible form of ANNs whose computational unit is a spiking neuron. Therefore, it's input and output are both in the form of spike events in time. Until recently, using backpropagation for SNNs has been quite challenging, especially for deep architectures. ANN to SNN conversion strategies\cite{Hunsberger2015,Esser2016,Liu2017,Diehl2015,Diehl2016,Rueckauer2017}
were the only available options to configure practical SNN systems. These methods, although effective, require a substantially large amount of time steps to produce a reliable output. In addition, they cannot be used to process event-based data directly as the data is already in the form of spike which cannot be handled natively by an ANN.

With recent efforts like \cite{Wu2019,Neftci2019,Lee2020,Shrestha2018,Jin2018,Panda2020}, it is now possible to train relatively deep SNNs (by SNN standards) using backpropagation. 
In these works, the loss measure is usually a mean-square error of the spike count at the output or a loss measure of the internal state, i.e. membrane potential of the output neurons \cite{ijcai2019-189}, \cite{MaXY21-0}. In this paper, we propose spike-based negative log-likelihood losses which are better suited for classification tasks. We first propose an output probability estimate from the output spike-trains and then subsequently use it in the cross-entropy setting. We further derive the gradient of this spike-based loss formulation. The proposed loss methods are hyperparameter free which is a plus. We call this spike-based loss formulation \emph{spikemax}.

We have released\footnote{
    The code for spikemax loss is publicly available at: \url{https://github.com/lava-nc/lava-dl/blob/main/src/lava/lib/dl/slayer/loss.py}
}
the implementation of spikemax loss as an add-on to SLAYER-PyTorch \cite{Shrestha2018} repository which is an SNN backpropagation implementation in PyTorch.

We experimentally evaluate the effectiveness of our loss measure with other existing alternatives on three different neuromorphic benchmark problems: NMNIST \cite{Orchard2015}, DVS Gesture \cite{Amir2017}, and N-TIDIGITS18 \cite{Anumula2018}. We use these neuromorphic benchmark datasets for evaluation because the SNN can directly process the raw data. As a result, there is no contribution of input to spike conversion method and the performance variation is solely due to the SNN training method. We report state of the art, if not competitive, results on these benchmark problems. In addition we also analyze the inference latency and spike activity of the resulting network which translate to power consumption in the hardware implementation and evaluate the methods on based on their relative power profile.

The rest of the paper is organized as follows. We will briefly introduce the preliminaries of an SNN and its gradient-based training. Then, we will delve into spike-based losses. Here we will formally introduce our \emph{spikemax} loss and its variants \emph{spikemax$_g$} and \emph{spikemax$_s$}. We follow it with experiments on the aforementioned neuromorphic benchmark datasets and their analysis. Finally, we will present the concluding remarks.

\section{Background}

In this section, we will briefly introduce the preliminary concepts of an SNN and gradient based training of an SNN in the spiking domain.

\subsection{SNN Model}

An SNN is a biologically plausible form of an ANN. Succinctly put, an SNN is a special form of an ANN which uses a spiking neuron as its activation/computational unit. A spiking neuron is a mathematical abstraction of a biological neuron. As a result an SNN inherits the intrinsic property of information exchange in the form of events in time known as spikes.

There are many models of a spiking neuron such as 
Leaky Integrate and Fire~(LIF) \cite{Gerstner2002,Paugam-Moisy2011},
Adaptive Exponential Integrate and Fire~(AdEx) \cite{Brette2005},
Izhikevich \cite{Izhikevich2003},
Spike Response Model~(SRM) \cite{Gerstner1995},
Hodgkin Huxley \cite{Hodgkin1952}, etc.
which represents the behavior of a biological neuron with a varying degree of realism. In this paper, we focus on the simple yet versatile SRM model of a spiking neuron. A SRM model of a spiking neuron is completely defined by a spike response kernel, $\varepsilon(\cdot)$ which describes the temporal response of the neuron to an input spike; a refractory kernel, $\nu(\cdot)$, which describes the post-spike behavior of the neuron; and a neuron threshold, $\vartheta$, which describes how easily the neuron spikes. We use $\varepsilon(t) = \rfrac{t}{\tau_s}\exp(1-\rfrac{t}{\tau_s})\Theta(t)$ and $\nu(t) = -2\vartheta\,\rfrac{t}{\tau_r}\exp(1-\rfrac{t}{\tau_r})\Theta(t)$ as the spike response kernel and refractory kernels where $\tau_s$ and $\tau_r$ are their respective time constants.

Mathematically, for a layer $l$ with $N_l$ neurons and inbound spikes, $\vct s^{(l-1)}(t)$, through the synaptic weights $\mat W^{(l-1)} \in \mathbb{R}^{N_l \times N_{l-1}}$, the internal voltage or the membrane potential, of the neuron is described as
\begin{align}
    \vct u^{(l)}(t) &= \mat W^{(l-1)} \left(\varepsilon * \vct s^{(l-1)}\right)(t) + \left(\nu * \vct s^{(l)}\right)(t)
\intertext{and the output of the layer is described as }
    \vct s^{(l)}(t) &= f_s\left(\vct u^{(l)}(t)\right)
\end{align}
where $f_s(u(t))=\sum_{t_f} \delta(t-t_f)\ : \{t_f : u(t_f) \geq \vartheta\}$ is the spike function.

\subsection{Gradient Based Training of SNN}

Error backpropagation has been the workhorse for the current advancement in deep learning. One of the earliest attempts for error backpropagation in SNN was formulated in SpikeProp \cite{Bohte2002a}. The authors used event-based error backpropagation. Further developments in the work have been proposed in \cite{ShresthaSong2016a,Shrestha2016,xu2013supervised,Comsa2019}

On the other end of the gradient-based learning for SNN spectrum is dt-based error backpropagation. These approaches \cite{Wu2019,Neftci2019,Zenke2018,Lee2020,Shrestha2018,Jin2018} have been successful in recent times for training deep SNNs. There are two main obstacles to be tackled for dt-based error backpropagation in SNN. The first is the derivative of the spike function (it is also an issue in event-based backpropagation) and the second is the temporal dependencies of the signals.

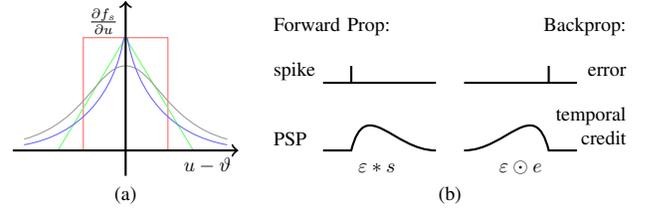
\begin{figure}
    \centering
    \begin{tikzpicture}[scale=\figscale, every node/.style={transform shape}]
    \begin{scope}
        \begin{scope}
            \draw[red!60] (-0.75, 0) |- (0, 2) -| (0.75, 0);
        \end{scope}
        \begin{scope}
            \draw[green!60] (-1.2, 0) -- (0, 2) -- (1.2, 0);
        \end{scope}
        \begin{scope}
            \draw[gray!80] (-1.8, 0.2) 
            .. controls (-0.7, 0.5) and (-0.5, 1.5) .. (0, 1.5)
            .. controls ( 0.5, 1.5) and ( 0.7, 0.5) .. (1.8, 0.2);
        \end{scope}
        \begin{scope}
            \draw[blue!60] ( 0.03, 2) to[out = -80, in = 170] ( 1.8, 0.1);
            \draw[blue!60] (-0.03, 2) to[out =-100, in = 10 ] (-1.8, 0.1);
        \end{scope}
        \begin{scope}
            \draw[->, thick] (-2, 0) -- (2, 0) node[anchor = north east] {$u - \vartheta$};
            \draw[->, thick] ( 0,-0.45) -- (0, 2.65) node[anchor = north east] {$\frac{\partial f_s}{\partial u}$};
        \end{scope}
        \node at (0, -0.8) {(a)};
    \end{scope}
    
    \begin{scope}[shift = {(3.5, 1.2)}]
        \begin{scope}
            \node[anchor = west] at (-1, 1) {Forward Prop:};
            \begin{scope}
                \node[anchor = west] at (-1, 0.2) {spike};
                \draw[thick] (0, 0) -- ++ (2, 0) (0.5, 0) -- ++(0, 0.3); 
            \end{scope}
            
            \begin{scope}[shift = {(0, -1.2)}]
                \node[anchor = west] at (-1, 0.2) {PSP};
                \draw[thick] (0, 0) -- (0.5, 0) .. controls (0.75, 1) and (1.25, 0) .. (2, 0);
            \end{scope}
            
            \begin{scope}[shift = {(0, -1.5)}]
                \node[anchor = west] at (0.5, 0) {$\varepsilon * s$};
            \end{scope}
        \end{scope}
        
        \begin{scope}[shift = {(2.5, 0)}]
            \node[anchor = east] at (3, 1) {Backprop:};
            \begin{scope}
                \node[anchor = east] at (3, 0.2) {error};
                \draw[thick] (0, 0) -- ++ (2, 0) (1.5, 0) -- ++(0, 0.3); 
            \end{scope}
            
            \begin{scope}[shift = {(0, -1.2)}]
                \node[anchor = east] at (3, 0.4) {\begin{tabular}{@{}r@{}}
                    temporal\\ credit
                \end{tabular}};
                \draw[thick] (2, 0) -- (1.5, 0) .. controls (1.25, 1) and (0.75, 0) .. (0, 0);
            \end{scope}
            
            \begin{scope}[shift = {(0, -1.5)}]
                \node[anchor = west] at (0.5, 0) {$\varepsilon \odot e$};
            \end{scope}
        \end{scope}
        \node at (2.25, -2) {(b)};
    \end{scope}
\end{tikzpicture}
    \caption{
        Error backpropagation in SNN
        (a) Spike function derivative.
        (b) Temporal credit assignment of an impulse error during backpropagation.
    }
    \label{fig:snnBackprop}
\end{figure}

Since, the spike generation mechanism in a spiking neuron is a discontinuous function, it's derivative does not exist in a mathematical sense. However, it is essential in the backpropagation chain for gradient propagation. This issue can be circumvented by using a proxy function for the spike function derivative. There are different possible representations of this proxy function such as spike escape rate function \cite{Shrestha2018}, or simply a linear function with finite voltage support \cite{Wu2019}, or a half-sigmoid function \cite{Zenke2018}. These proxy functions are best interpreted as a surrogate gradient function as described in \cite{Neftci2019}. Various forms of surrogate gradient functions are depicted in \reffig{fig:snnBackprop}~(a).

Another important, and often overlooked (often for computational reasons) aspect of dt-based error backpropagation in SNN. However, it is important because of the inherent temporal nature of a spiking neuron. An input spike at a point of time to a spiking neuron induces a post-synaptic response over a range of time. Therefore, it is necessary to compensate for this behavior during error backpropagation. As shown in Slayer \cite{Shrestha2018}, this forward temporal effect (represented by the convolution operation in time: $*$) must be compensated by similar temporal distribution of error signal back in time i.e. correlation operation ($\odot$). This is illustrated in \reffig{fig:snnBackprop}~(b).

In this work, we use SLAYER-PyTorch \cite{Shrestha2018} framework as the SNN training method as it is publicly available\footnote{
    SLAYER-PyTorch is publicly available at: \url{https://github.com/bamsumit/slayerPytorch}
} and has proven to be effective for training SNNs.

\section{Spike Based Losses}

The training cost of the SNN, given a target spike train $\hat{\vct s}(t)$ is usually formulated as
\begin{align}
    \mathcal L = \int_0^T l(\vct s^{(n_l)}(t), \hat{\vct s}(t))\,\text dt
\end{align}
where $n_l$ is the last/output layer and $t\in[0, T]$ is the simulation time interval. In spike regression problem, the target spike train is usually known. In this case, with $l(t) = \left\{\varepsilon * \left(\vct s^{(n_l)} -\hat{\vct s}\right)(t)\right\}^2$, the cost can be formulated as a van-Rossum distance \cite{dauwels2008similarity}. 

However, the target spike train is not known for classification problems. Typically, in SNN based classification, we want the true output neuron to spike the most number of times. One of the prevalent strategies is to maximize the membrane potential of the true output neuron \cite{Lee2020,Kaiser2020}, usually at the end of simulation time. One would have to wait till the end of the simulation to get the classification output. In addition, in a real neuromorphic hardware, the membrane potential is not visible. Although maximizing membrane potential usually results in more spike count, however, it does not guarantee it always.

Another common strategy is to directly maximize the spike-count or spike-rate \cite{Wu2019,Shrestha2018}. The \emph{spike-rate} loss can be formulated as
\begin{align}
    \mathcal L = \left\{\frac{1}{T}\int_0^T \vct s^{(n_l)}(t)\,\text dt - \hat{\vct r}\right\}^2
\end{align}
where $\hat{\vct r}$ is the desired spike rate at the output.

Next, we will formulate negative log-likelihood losses based on the probability interpretation of spikes. This is the main theoretical contribution of the paper.

\subsection{Probability Interpretation of Spikes}
Each spike event at the output is like a vote for the output being that particular class. The count of the spike, thus, represents the totality of the votes, which can be defined as
\begin{align}
    c_i(t) = \int_{t-W}^t s_i(\tau)\,\text d\tau
\end{align}
where $W$ is the spike count estimate window and $i\in\{0, 1, \cdots, N_{n_l}-1\}$ is the neuron index. Then the probability estimate of the output at time $t$ is
\begin{align}
    p_i(t) = \frac{c_i(t)}{\sum_{i=0}^{N_{n_l}-1}c_i(t)}
\end{align}

One can make the length of the sliding window as large as the simulation interval. In that case, the global probability estimate is
\begin{align}
    p_i = \frac{c_i}{\sum_{i=0}^{N_{n_l}-1}c_i},
    \qquad\text{where }
    c_i = \int_0^T s_i(\tau)\,\text d\tau
\end{align}
Note the absence of time dependency disambiguates between running probability estimate and global probability estimate.

\subsection{Spikemax Losses}
With the probability estimate from the spike output and one-hot target output $\hat{y}_i(t)=\hat{y}_i$, we formulate the negative log-likelihood loss as follows:
\begin{align}
    \mathcal L = \frac{1}{T}\int_0^T l(t)\,\text dt\ ,\ \  
    l(t) = -\sum_i \hat{y}_i(t) \log\big(p_i(t)\big)
\end{align}
We term this formulation of loss as \emph{spikemax}. The gradients with respect to spikemax loss are
\begin{align}
    \frac{\partial\mathcal L}{\partial p_i(t)} 
    &= \frac{\partial\mathcal L}{\partial l(t)} \frac{\partial l(t)}{\partial p_i(t)}
    = -\frac{\hat{y}_i(t)}{p_i(t)}\\
    \frac{\partial p_k(t)}{\partial c_i(t)}
    &= \begin{cases}
        \frac{1-p_i(t)}{\sum_i c_i(t)} &\text{if } i=k\\
        -\frac{p_k(t)}{\sum_i c_i(t)}  &\text{otherwise}
    \end{cases}\\
    \frac{\partial\mathcal L}{\partial s_i(t)}
    &= \left(\frac{\partial\mathcal L}{\partial p_i(t)} \frac{\partial p_i(t)}{\partial c_i(t)}
    + \sum_{k\neq i} \frac{\partial\mathcal L}{\partial p_k(t)} \frac{\partial p_k(t)}{\partial c_i(t)}  
    \right)\frac{\partial c_i(t)}{\partial s_i(t)} \nonumber\\
    &= \frac{p_i(t) - \hat{y}_i(t)}{c_i(t) / W}.
\end{align}
Note, for numerical stability, an infinitesimal count can be added. In addition, using the global probability estimate, we define the \emph{spikemax$_g$} loss and derive it's gradient as
\begin{align}
    \mathcal L = -\sum_i\hat{y} \log(p_i)
    ,\qquad
    \frac{\partial\mathcal L}{\partial s_i(t)}
    &= \frac{p_i - \hat{y}_i}{c_i/T}.
\end{align}

It is also possible to use softmax probability estimate from spike count and use negative log-likelihood loss.
\begin{align}
    p^\text{s}_i &= \frac{\exp(c_i)}{\sum_{i=0}^{N_{n_l}-1}\exp(c_i)}\\
    \mathcal L &= -\sum_i\hat{y} \log(p^\text{s}_i)
    ,\qquad
    \frac{\partial\mathcal L}{\partial s_i(t)}
    &= p^\text{s}_i - \hat{y}_i
\end{align} 
We will simply call it \emph{spikemax$_s$} in the rest of this paper.

With losses and their respective gradient formulation derived, we can incorporate them in the SLAYER-PyTorch computational graph to train an SNN. The advantage of the proposed loss methods in contrast to the common spike-rate loss formulation is that spikemax losses are parameter free i.e. there is no need to tune the output spike rates manually. In addition, spikemax loss optimizes the spiking output for producing continuous classification output using a sliding window, therefore, is more suitable for use in a real-time system.
\section{Experiments and Results}


In this section, we will compare the performance of spikemax losses with spike-rate loss on three different neuromorphic classification tasks. We use SLAYER-PyTorch \cite{Shrestha2018} as our SNN backpropagation framework. Our loss methods are implemented on top of it.

All the results presented in this paper are averaged over 5 different independent trials. Same random seeds were used for each of the loss methods for a fair comparison. We will follow the shorthand notation in \cite{Shrestha2018} to represent the architecture: layers are separated by \verb~-~, spatial dimensions are separated by \verb~x~, an $N\times N$ convolution filter with $K$ channels is represented by \verb~KcN~, an $N\times N$ aggregate pooling filter is represented by \verb~Na~ and a dense layer with N neurons is represented by the number itself.  Note that the convolution and dense layer neurons also have trainable axonal delays \cite{axonaldelay}. In all our experiments, we use neuron threshold $\vartheta =10$~mv, and sampling time of $1$~ms. The spike response time constant, $\tau_s$, and refractory time constant, $\tau_r$, were optimized for each of the datasets. We consider the datasets with inputs as spike events which are then fed directly to the SNN. This also eliminates spike encoding out of the processing pipeline and hence we can focus on SNN training only.

We will compare the results based on the overall accuracy obtained. We will also see how the methods compare in terms of inference latency. In a real system, the network that can classify with reliable accuracy faster does not need to look at the entire duration of the input, therefore, can result in a power efficient inference system. We compare the network's classification accuracy versus the time-length of the input sequence to evaluate the network's latency for classification. Lower classification latency means that we do not need to process the rest of the inputs which saves the energy consumption of the end system. In addition, we will compare the networks in terms of average spike count. Since we are not implementing the system in a hardware, spike count is a good measure of the relative inference energy of the network \cite{Lee2020}. More spike activity usually means more energy is consumed by the neuromorphic hardware during inference. Ideally, one would want higher accuracy, earlier inference and low spike count.

\begin{table}
\small
	\centering
	\caption{Benchmark Classification Results}
	\label{tbl:results}
	\begin{tabular}{|c|l|r|c|}
		\cline{2-4}
		\multicolumn{1}{c|}{}& \multicolumn{1}{c|}{\bf Method} & \multicolumn{1}{c|}{\bf Params} & \textbf{Accuracy}
		\\ \hline
		\multirow{7}{*}{\rotatebox{90}{NMNIST}}
		& Lee et al. \cite{Lee2016} & 1,857,600 & $98.66\%$ \\
		& Wu et al. \cite{Wu2019}	& 17,664,256 & $99.53\%$ \\
		
		& Spike-based BP. \cite{fang2021incorporating}	& 68,537,760 & $\bf99.61\%$ \\
				
		& spike-rate	&  2,171,728 & $99.33\pm 0.03\%$ \\
		& spikemax		&  2,171,728 & $99.27\pm 0.02\%$ \\
		& spikemax$_g$	&  2,171,728 & $99.33\pm 0.04\%$ \\
		& spikemax$_s$	&  2,171,728 & $99.26\pm 0.06\%$ \\
		\hline
		\multirow{8}{*}{\rotatebox{90}{DVS Gesture}}
		& TrueNorth \cite{Amir2017}
						& 1,992,476 & $91.77 (94.59)\%$ \\
		& DECOLLE \cite{Kaiser2020}	 
		                & 1,245,696 & $95.54\%$ \\
		& Ghosh et al. \cite{Ghosh2019}$^\dagger$ 
		                & 2,119,080 & $95.94\%$ \\
		& Spike-based BP. \cite{fang2021incorporating}
		                & 6,798,292 & $\bf97.57\%$ \\

		& spike-rate	& 1,068,368 & $96.21\pm 0.63\%$ \\
		& spikemax		& 1,068,368 & $95.83\pm 0.48\%$ \\
		& spikemax$_g$	& 1,068,368 & $95.53\pm 0.37\%$ \\
		& spikemax$_s$	& 1,068,368 & $95.15\pm 0.65\%$ \\
		\hline
		\multirow{8}{*}{\rotatebox{90}{N-TDIDIGITS18}}
		& GRU-RNN \cite{Anumula2018}$^\dagger$
						& 109,200 & $90.90\%$ \\
		& Phased-LSTM \cite{Anumula2018}$^\dagger$
						& 610,500 & $91.25\%$ \\
		& ST-RSBP \cite{Zhang2019}
						& 351,241 & $93.63\pm 0.27\%$ \\
		& spike-rate	&  84,736 & $\bf94.19\pm 0.18\%$ \\
		& spikemax		&  84,736 & $93.21\pm 0.32\%$ \\
		& spikemax$_g$	&  84,736 & $93.01\pm 0.38\%$ \\
		& spikemax$_s$	&  84,736 & $92.43\pm 0.25\%$ \\
		\hline
		\multicolumn{3}{l}{\footnotesize{$^\dagger$ Non SNN implementation.}}
	\end{tabular}
	\vspace{-0.5cm}
\end{table}

\subsection{NMNIST digit classification}

		


NMNIST dataset \cite{Orchard2015} is the neuromorphic version of standard MNIST images. The images are converted into spikes using a Dynamic Vision Sensor~(DVS) moving on a pan-tilt unit in a three-saccadic motion, each lasting $100$ ms long. Resulting event-data is $34\times 34$ pixels, with ``on'' and ``off'' spike-events. The events last for $300$ ms per sample. In our experiments, we do not stabilize the NMNIST data to compensate the saccadic movement. Raw spike data is used, without any processing. The train and test split is the same as standard MNIST: 60,000 training samples and 10,000 testing samples. 

We use a spiking CNN architecture with the following specification: \verb~34x34x2-16c5-2a-32c3-2a-64c3~ \verb~-512-10~. The neuron time constants are $(\tau_s, \tau_r) = (1, 1)$~ms, the target spike rate are $(\text{True},\text{False}) = (0.2,0.04)$ for spike-rate loss, and $W=30$ for spikemax loss. The overall classification results for our loss methods are listed in \reftab{tbl:results} with other reported benchmarks. The overall accuracy for spike-rate, spikemax$_s$, spikemax$_g$ and spikemax loss are very similar at around $99.3\%$. The result is lower than the best reported accuracy of $99.61\%$ \cite{fang2021incorporating} on NMNSIT. However, the network we use is significantly smaller ($31\times$ less parameter).

\begin{figure}
	\centering
	\includegraphics[scale=\figscale]{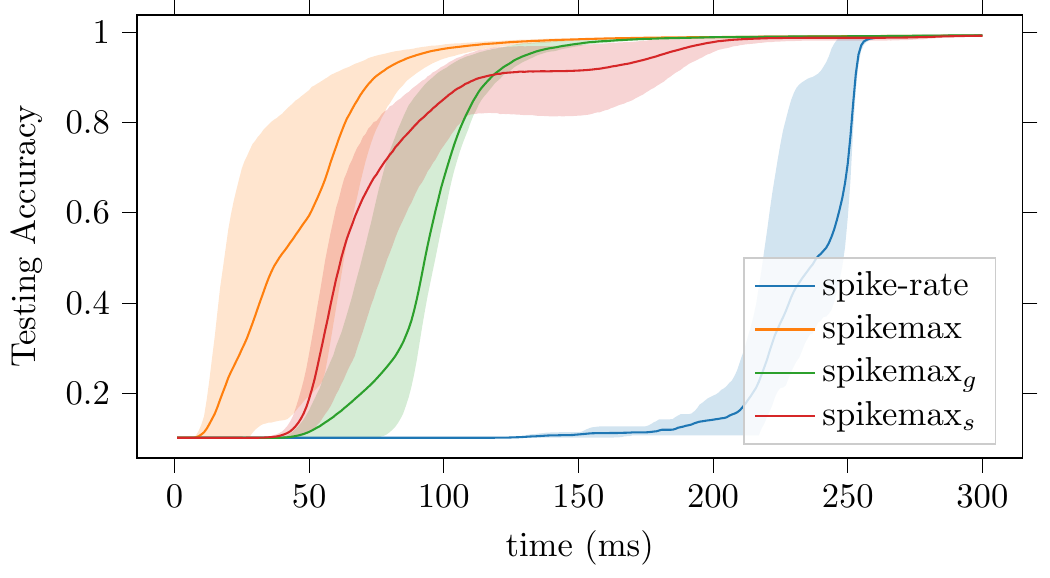}
	\caption{Testing accuracy over inference runtime for NMNIST classification.}
	\label{fig:nmnistAcc}
\end{figure}

\begin{figure}
	\centering
	\includegraphics[scale=\figscale]{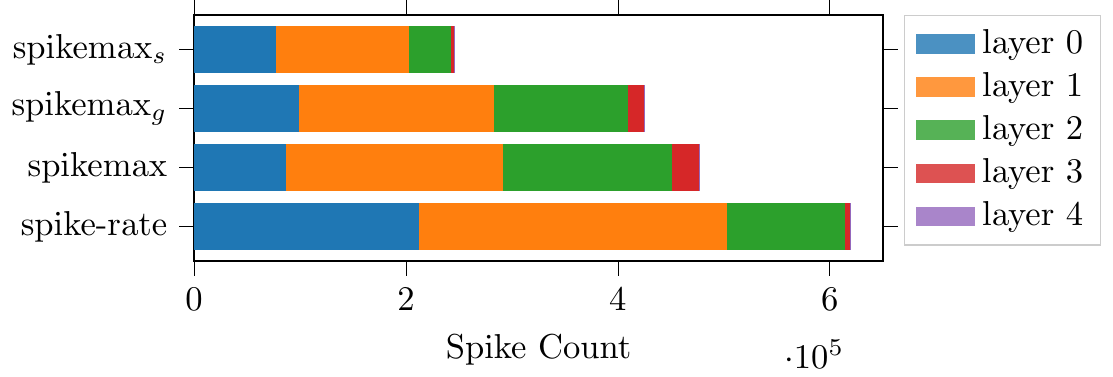}
	\caption{Spike count distribution per layer for NMNIST classification.}
	\label{fig:nmnistCnt}
\end{figure}

\reffig{fig:nmnistAcc} shows the plot of accuracy and inference latency. Spikemax networks clearly show the minimum inference latency ($\approx 70$~ms) among all the methods whereas spike-rate networks demonstrate the slowest inference latency ($\approx 250$~ms). The networks trained with negative log-likelihood losses (spikemax$_s$, spikemax and spikemax$_g$) clearly show faster inference times compared to spike-rate networks.

The plot of average spike count per layer is shown in \reffig{fig:nmnistCnt}. The networks trained with spike-rate spike the most whereas spikemax$_s$ networks demonstrate minimum spike.

\subsection{DVS Gesture classification}


The DVS Gesture \cite{Amir2017} is a public dataset by IBM Research. It is a neuromorphic dataset that consists of 29 different individuals performing 11 hand gestures (clapping, drumming, hand-wave, etc.) under three different lighting environments captured using a DVS camera. The standard train-test split of first 23 subjects for training and the last 6 subjects for testing is used. The data is $128\times 128$ pixels wide with ``on'' and ``off'' polarity. We train on randomly sampled $300$~ms long sequence and test on first $1.5$~s event sequence.

The network architecture is \verb~128x128x2-4a-16c5~ \verb~-2a-32c3-2a-512-11~ . The neuron time constants are $(\tau_s, \tau_r) = (5, 5)$~ms, the target spike rate are $(\text{True},\text{False}) = (0.35,0.07)$ for spike-rate loss, and $W = 35$ for spikemax loss. The results are listed in \reftab{tbl:results}. We report the best accuracy of $96.97\%$ using spikerate loss. spikemax$_s$, spikemax and spikemax$_g$ results are also very good: better than the results using non-spiking CNN \cite{Ghosh2019} and other SNN based approaches. Compared with the latest best-reported result \cite{fang2021incorporating}, we only use $6\times$ less parameter. 

\begin{figure}
	\centering
	\includegraphics[scale=\figscale]{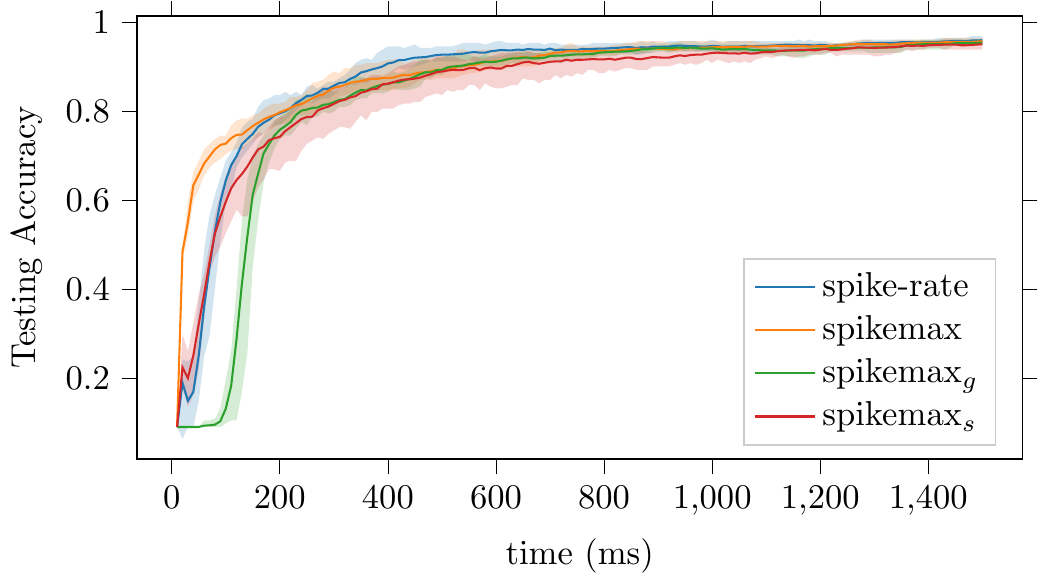}
	\caption{Testing accuracy over inference runtime for DVS Gesture classification.}
	\label{fig:dvsGestureAcc}
\end{figure}

\begin{figure}
	\centering
	\includegraphics[scale=\figscale]{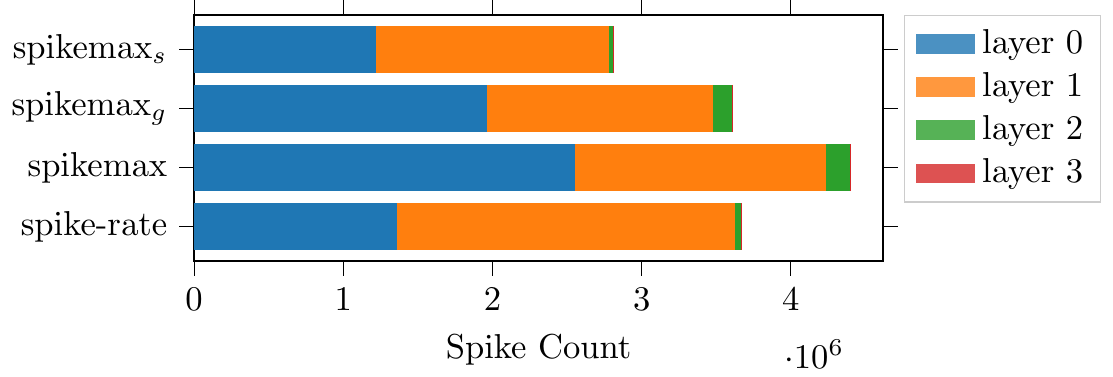}
	\caption{Spike count distribution per layer for DVS Gesture classification.}
	\label{fig:dvsGestureCnt}
\end{figure}

\reffig{fig:dvsGestureAcc} shows the plot of accuracy versus inference time for all the loss methods. We can observe that spikemax loss is able to reliably predict the classification output as early as approx. $50$ ms. The networks trained using spike-rate and spikemax$_s$ show inference latency of approx. $100$~ms and the networks trained using spikemax show inference latency of approx. $150$~ms. Past $200$~ms, the inference accuracy of all the networks are similar and saturate after $700$~ms.

The average spike count distribution of each layer of the networks is shown in \reffig{fig:dvsGestureCnt} for the final inference time of $1500$~ms. Spikemax$_g$ networks demonstrate the least spike activity among all, therefore, are relatively energy efficient. However, these networks require a longer inference time.

\subsection{N-TIDIGITS18 audio classification}




N-TIDIGITS18 \cite{Anumula2018} is the neuromorphic version of the TIDIGITS \cite{Leonard1993} audio classification dataset. The TIDIGITS audio signals were converted into a $64$ channel spiking events using a silicon cochlea sensor: CochleaAMS1b \cite{Liu2013}. The output classes are digit utterances from ``zero'' to ``nine'' and ``oh''. We use the standard train-test split of the dataset as used in \cite{Anumula2018}.

We use a fully connected architecture, \verb~64-256-256-11~, with neuron time constants, $(\tau_s, \tau_r) = (5, 5)$~ms, $(\text{True},\text{False})=(0.2, 0.02)$, and $W=40$, in our experiments. The classification results are tabluated in \reftab{tbl:results}. We achieve the best performance of $94.45\%$ using spike-rate loss, better than the recurrent SNN architecture \cite{Zhang2019} with $4\times$ less parameters. Networks trained with spikemax also show good classification performance. Note that we achieve better results than the conventional LSTM network \cite{Anumula2018} for all the loss methods with $7\times$ less network parameters. 

\begin{figure}
	\centering
	\includegraphics[scale=\figscale]{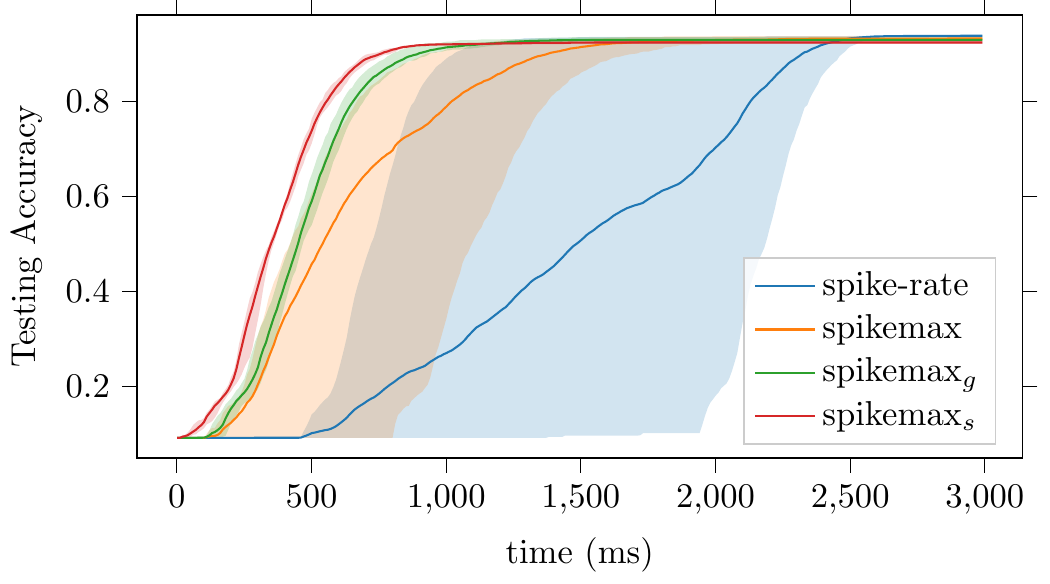}
	\caption{Testing accuracy over inference runtime for N-TIDIGITS18 classification.}
	\label{fig:ntidigitsAcc}
\end{figure}

\begin{figure}
	\centering
	\includegraphics[scale=\figscale]{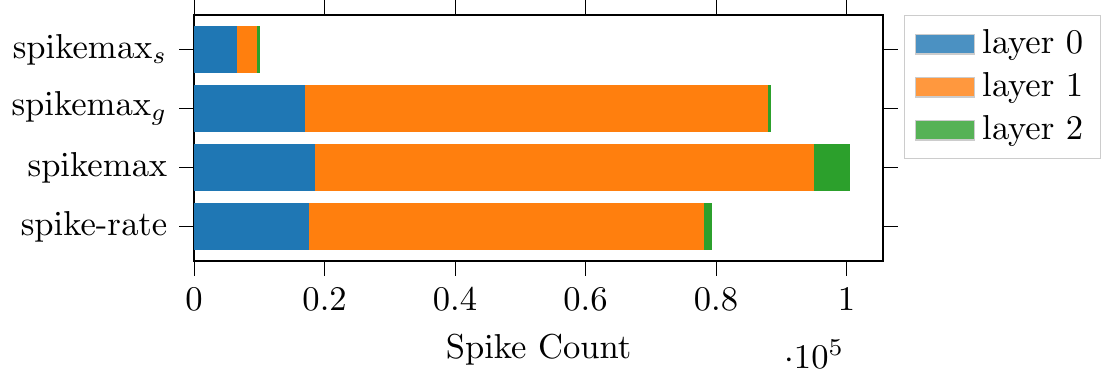}
	\caption{Spike count distribution per layer for N-TIDIGITS18 classification.}
	\label{fig:ntidigitsCnt}
\end{figure}

Accuracy versus inference time is plotted in \reffig{fig:ntidigitsAcc}. It is clear that the networks trained with negative log-likelihood based losses (spikemax$_s$, spikemax and spikemax$_g$) are able to classify much earlier than the networks trained with spke-rate: approx. $600$~ms compared to approx. $2000$~ms. 

In \reffig{fig:ntidigitsCnt}, the average spike count per layer for the trained networks is plotted. The spikemax$_s$ networks demonstrate 
a remarkably low spike count. They also show earlier inference amongst all other networks.

\section{Discussion}
In this paper, we have proposed spike-based negative log-likelihood based losses suitable to train an SNN for classification tasks. We demonstrate state of the art, if not competitive, classification performances on neuromorphic audio and video datasets using these loss methods. We focus on purely neuromorphic datasets so that we exclude the effectiveness of spike-encoding process in the comparison and work directly on the raw spike data.

We also compare the resulting networks in terms of their inference time as well as the spike activity of the network. These measures are a proxy for the energy consumed per inference in a neuromorphic hardware. From the results, we can clearly see that the proposed losses result in networks that start to produce usable inference results earlier. Spikemax$_s$ loss, in particular, demonstrated considerably low spike activity in the trained networks in two of the three benchmarks. 

In addition, we also demonstrate very good classification performances, using networks with fewer learnable parameters. Particularly in N-TIDIGITS18 audio classification tasks, our networks were able to outperform spiking recurrent networks as well as conventional LSTM and GRU networks. This performance is not solely due to the loss method. 

In conclusion, we show a way to use negative log-likelihood loss in the spiking domain by estimating the probability confidence from spikes. We see that these losses demonstrate improvements in spike-based classification tasks, especially from the energy perspective. The idea of probability estimate can also be extended for other spike-based learning tasks. We will consider these avenues in the future.

\section{Acknowledgments}
This research is partially supported by Programmatic grant no. A1687b0033 from the Singapore government’s Research, Innovation and Enterprise 2020 plan (Advanced Manufacturing and Engineering domain) and the Flemish Government under the "Onderzoeksprogramma Artificiele Intelligentie (AI) Vlaanderen".


\bibliographystyle{IEEEbib}
\bibliography{bibliography}

\end{document}